\pdfoutput=1

\documentclass[11pt]{article}

\usepackage{acl}

\usepackage{times}
\usepackage{latexsym}

\usepackage[T1]{fontenc}

\usepackage[utf8]{inputenc}

\usepackage{microtype}

\usepackage{xcolor}
\definecolor{lightgray}{gray}{0.95}

\usepackage{listings}
\newcommand{\name}[1]{\textsc{QUDSelect}}
\newcommand{\llama}[1]{LLaMA2}
\newcommand{\mistral}[1]{Mistral}

\title{\name{}: Selective Decoding for Questions Under Discussion Parsing}

\usepackage{graphicx}
\usepackage{multirow}
\usepackage{makecell}
\usepackage{booktabs}
\usepackage{enumitem}
\usepackage{amssymb}
\usepackage{xcolor}
\usepackage{color}
\usepackage{colortbl}
\usepackage{multirow}
\definecolor{dark-green}{rgb}{0.31, 0.47, 0.26}
\definecolor{dark-red}{rgb}{0.81, 0.09, 0.13}
\definecolor{airforceblue}{rgb}{0.36, 0.54, 0.66}

\usepackage{soul}
\usepackage{amsmath}

\setuldepth{Berlin}
\usepackage{tabularx}
\newcolumntype{x}[1]{>{\arraybackslash\hspace{0pt}}m{#1}}

\usepackage{pifont}
\newcommand{\cmark}{\color{dark-green}{\ding{51}}}
\newcommand{\xmark}{\color{dark-red}{\ding{55}}}

\usepackage{todonotes}

\usepackage{cleveref}
\crefformat{section}{\S#2#1#3}
\crefformat{subsection}{\S#2#1#3}
\crefformat{subsubsection}{\S#2#1#3}
\crefrangeformat{section}{\S\S#3#1#4 to~#5#2#6}
\crefmultiformat{section}{\S\S#2#1#3}{ and~#2#1#3}{, #2#1#3}{ and~#2#1#3}
\crefmultiformat{subsection}{\S\S#2#1#3}{ and~#2#1#3}{, #2#1#3}{ and~#2#1#3}
\Crefformat{figure}{#2Fig. ~#1#3}
\Crefformat{table}{#2Tab.~#1#3}
\Crefformat{appendix}{Appx.~\S#2#1#3}
\crefformat{algorithm}{Alg.~#2#1#3}
\Crefformat{equation}{Eq.~#2#1#3}
\author{
  Ashima Suvarna$^{\heartsuit}$\thanks{\ \  Equal contribution.}\, \, Xiao Liu$^{\diamondsuit*}$\, \, Tanmay Parekh$^{\heartsuit}$\, \, Kai-Wei Chang$^{\heartsuit}$\, \, Nanyun Peng$^{\heartsuit}$\\
  $^{\heartsuit}$ Computer Science Department, University of California, Los Angeles \\
  $^{\diamondsuit}$ Wangxuan Institute of Computer Technology, Peking University \\
  \texttt{ \{asuvarna31,tparekh,kwchang,violetpeng\}@cs.ucla.edu}\\ 
  \texttt{lxlisa@pku.edu.cn} 
}

\begin{document}
\maketitle

\begin{abstract}

Question Under Discussion (QUD) is a discourse framework that uses implicit questions to reveal discourse relationships between sentences. In QUD parsing, each sentence is viewed as an answer to a question triggered by an anchor sentence in prior context. The resulting QUD structure is required to conform to several \emph{theoretical criteria} like answer compatibility(how well the question is answered), making QUD parsing a challenging task. 
Previous works construct QUD parsers in a pipelined manner (i.e. detect the trigger sentence in context and then generate the question). However, these parsers lack a holistic view of the task and can hardly satisfy all the criteria.
In this work, we introduce \textbf{\name{}}, a joint-training framework that selectively decodes the QUD dependency structures considering the QUD criteria. Using instruction-tuning, we train models to simultaneously  predict the anchor sentence and generate the associated question. To explicitly incorporate the criteria, we adopt a selective decoding strategy of sampling multiple QUD candidates during inference, followed by selecting the best one with criteria scorers. Our method outperforms the state-of-the-art baseline models by $9\%$ in human evaluation and $4\%$ in automatic evaluation, demonstrating the effectiveness of our framework.\footnote{We plan to release the code and models soon.}

\end{abstract}
\section{Introduction}
\label{sec:intro}

\begin{figure}[t]
    \centering
    \includegraphics[width=\linewidth]{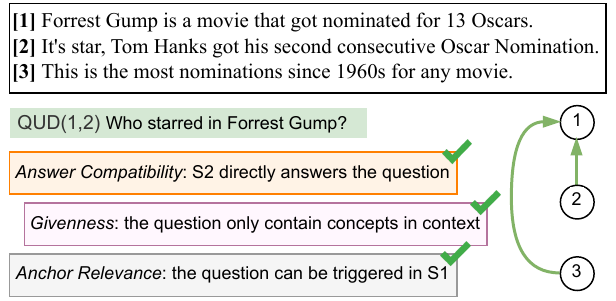}
    \caption{An article snippet along with the associated QUD dependency structure. Each edge from $s_i$ to $s_j$ with attribute $q$ indicates sentence $s_j$ anchors the question $q$, and sentence $s_i$ answers the question $q$.}
    \label{fig:task}
\end{figure}

Discourse structure describes the relationships between different sentences of an article or conversation. The ability to understand discourse structure is crucial for natural language processing tasks such as text summarization \cite{durrett2016learning}, conditional generation \cite{narayan-etal-2023-conditional}, and narrative understanding \cite{xu2024graph}. Recent works have adapted the Question Under Discussion (QUD) framework to analyze discourse structures~\cite{benz2017questions,riester2021combined}.
In the QUD framework~\cite{van1995discourse,roberts2012information}, the relationships between sentences in an article are characterized by (implicit) free-form questions. Each question is evoked by an anchor sentence in prior context, and answered by an answer sentence in the subsequent content. 
For instance, in Figure \ref{fig:task}, the relationship between sentence 3 (referred to as $s_3$) and the previous context is that $s_3$ answers the question ``Which movie has the most Oscar nominations?'' evoked by the anchor sentence $s_1$. 

\begin{figure*}[ht]
    \centering
    
    \includegraphics[scale=0.37]{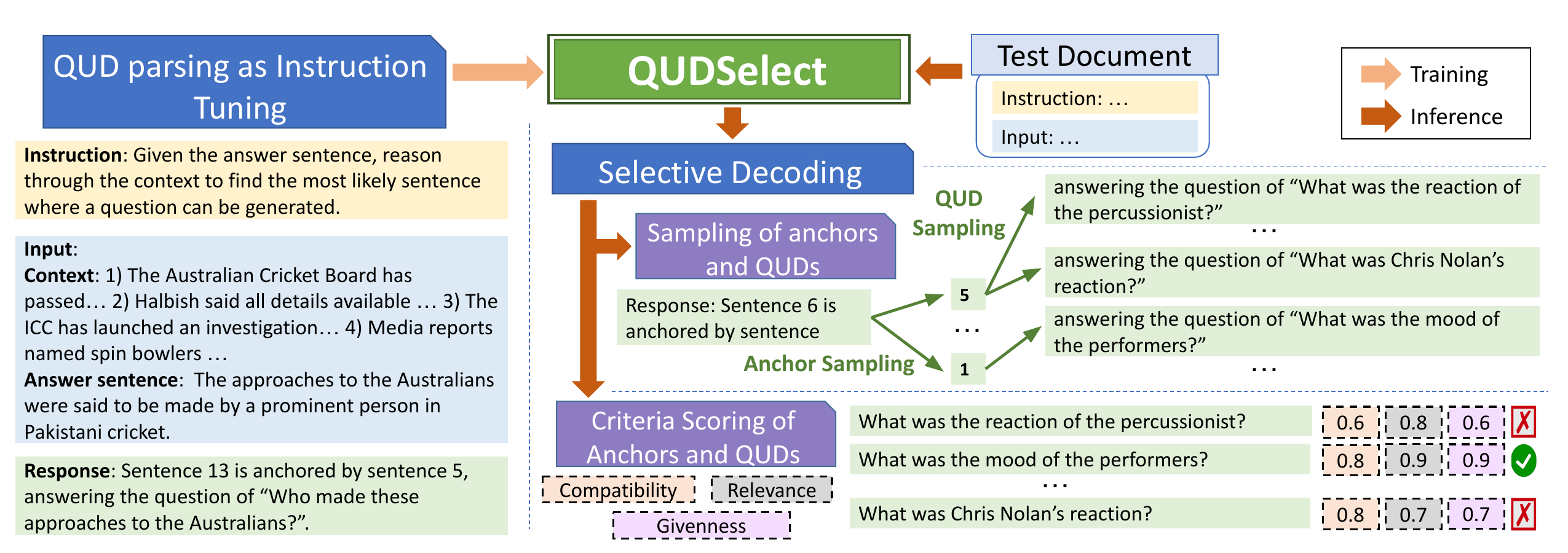}
    \caption{Overview of our \name{} framework.}
    \label{fig:method}
\end{figure*}

The QUD structures involve contextually-grounded questions that adhere to three theoretical criteria \cite{de2018qud,wu-etal-2023-qudeval, riester2018annotation}: a) \emph{answer compatibility}: the question must be answerable by the answer sentence in the discourse, like $s_2$ directly answers the question ``Who starred in Forrest Gump?'' in Figure~\ref{fig:task}; b) \emph{givenness}: the question should only contain concepts that are accessible to the reader from prior context or common knowledge, like ``Forrest Gump'' in the question; and c) \emph{anchor relevance}: the question should be relevant to the anchor sentence, e.g., the aforementioned question can be triggered in $s_1$. 

Previous works on QUD parsing break down the task into two steps: anchor selection and question generation. \citet{de2020towards} develop a rule-based method for the question generation step, \citet{ko-etal-2023-discourse} train task-specific models for each step, while \citet{wu-etal-2023-qudeval} prompt large language models (LLMs) in a stepwise manner. However, these approaches lack a holistic view of the task, causing the predicted QUDs to often fail to satisfy all the criteria. 
For instance, GPT-4 fails to generate questions that are fully grounded on the anchor sentence in $50\%$ of the cases.\footnote{This is observed from the human annotations in the QUD evaluation dataset \textsc{QUDeval}~\cite{wu-etal-2023-qudeval}.}

To address these challenges, we propose \name{}, a joint-training framework that selectively decodes QUD structures by incorporating the criteria, as shown in Figure~\ref{fig:method}. Specifically, we instruction-tune models to jointly predict the anchor sentence and the corresponding question given an answer sentence (e.g., $s_{13}$) and prior context (e.g., $s_1, \ldots, s_{12}$ of the article). We propose selective decoding where we sample multiple anchor and question pairs, score them using criteria scorers, and finally, select the best scored pair.

Experiments conducted on the DCQA~\cite{ko-etal-2022-discourse} dataset show that \name{} outperforms baselines by \textasciitilde$9\%$ on average in human evaluation. To reduce resource and cost-intensive expert evaluation, we develop automatic evaluators trained on human annotations, and conduct a larger-scale automatic evaluation. The automatic evaluation results show that \name{} achieves around a \textasciitilde$4\%$ improvement over the selected baselines. Further analyses reveal that the performance could be further improved with more selected candidates.

\begin{table*}[ht]
    \centering
    \small
    \setlength{\tabcolsep}{2.6pt}
    \resizebox{\linewidth}{!}{%
    \begin{tabular}{l|ccc|ccc|ccc|c}
        \toprule
        \multirow{2}{*}{\textbf{Model}} & \multicolumn{3}{c|}{\textbf{Answer Compatibility}} & \multicolumn{3}{c|}{\textbf{Givenness}} & \multicolumn{3}{c|}{\textbf{Anchor Relevance}} & \multirow{2}{*}{\ \textbf{Avg. ($\uparrow$)}\ } \\
        & \textbf{Dir. ($\uparrow$)} & \textbf{Unfocus. } & \textbf{No Ans.($\downarrow$)} & \textbf{No New ($\uparrow$)} & \textbf{Ans. leak. ($\downarrow$)} & \textbf{Hall. ($\downarrow$)}        &  \textbf{Fully G. ($\uparrow$)} & \textbf{Partial. G. } & \textbf{No G. ($\downarrow$)} & \\
        \midrule
        \multicolumn{11}{c}{\textsc{Automatic Evaluation}} \\
        \midrule
        Pipeline & 68.2 & 4.5 & 27.3 & 83.7 &10.0 &	6.3 & 63.6 & 0.0 & 36.4 & 71.8 \\
        \llama{}-7B & 67.4 & 12.9 & 19.7 & 88.3	& 6.7 & 5.0  & 52.7 & 17.7 & 29.6 & 69.5\\
        + \name{} & 70.4 &	8.2	& 21.4 & \textbf{91.8} & \textbf{6.0}	& \textbf{2.2} &  61.0 &	12.4 & 26.6 & 74.4 \\
        \mistral{}-7B & 71.4 & 8.7 & 19.9 & 89.3 & \textbf{6.0} &	4.7 & 58.0 & 15.9 & 26.1 & 72.9 \\
        + \name{} & \underline{74.1} & 9.0 & \underline{16.9} & 86.5 & 7.2&	6.2 & \textbf{68.3} & 11.0 & \underline{20.7} & \underline{76.3}\\
        GPT-4 & \textbf{92.7} &	3.3	& \textbf{4.0} & 78.7 &	18.9 &	2.4 & 51.9	 & 32.0 &	16.1 & 74.4 \\ 
        + \name{} & 90.0 &	4.1	&5.9	&80.0	&15.0	&5.0 &62.5 &	21.4	& \textbf{16.0} & \textbf{77.5} \\
        \midrule
        \multicolumn{11}{c}{\textsc{Human Evaluation}} \\
        \midrule
        Pipeline &52.5	&15.0	&32.5	&53.8	&28.7	&17.5  & 50.0	& 32.5	&17.5 & 52.1	\\
        Mistral-7B & 67.0 &	15.4 &	17.6 & 60.3	& 23.6	& 16.1 & 58.6	& 29.0	& 12.4 & 62.0	\\
        + \name{} &	\textbf{67.1} &	20.0 & \textbf{12.9} & \textbf{77.6} &	\textbf{20.0} &	\textbf{2.4} & \textbf{68.2}	& 24.7 &	\textbf{7.1} & \textbf{71.0} \\
        \bottomrule
    \end{tabular}%
    }
    \caption{Automatic and human evaluation results. Numbers are in percentages (\%). Best results are in bold, and the best results of open-source models (if not the best overall) are underlined. Avg. indicates the average ratio of ideal QUDs (the first option of each criterion). We abbreviate Direct Answer as Dir. Ans., Indirect Answer as Indir. Ans., Answer Leakeage as Ans. Leak., Hallucination as Hall., and Grounded as G.}
    \label{tab:auto_eval}
\end{table*}

    

\section{Related Work}
\label{sec:related}
QUD is a linguistic framework that analyzes discourse and pragmatics by viewing each sentence as an answer to an implicit question triggered in prior context \cite{van1995discourse, roberts2012information, benz2017questions}. While theoretical discussions around QUDs relied on constructed examples, \citet{riester2019constructing} introduced an annotation framework for reconstructing QUDs from data. \citet{westera-etal-2020-ted}, \citet{ko-etal-2022-discourse} and \citet{hesse2020annotating} annotated Ted-talk transcripts and news articles respectively in an expectation-driven manner, where questions are triggered while reading (i.e., unseen discourse progression) while \citet{de2018qud} annotated two interview transcripts with full, hierarchical questions. 

Recent works have begun adapting QUD for automatic discourse parsing \cite{ko-etal-2022-discourse, ko-etal-2023-discourse, wu-etal-2023-qudeval}, narrative graph construction \cite{xu2024graph} and decontextualization of scientific documents \cite{newman2023question}. \citet{ko-etal-2023-discourse} introduced a QUD parser trained on DCQA \cite{ko-etal-2022-discourse} that consists of an anchor selection and a question generation pipeline. \citet{wu-etal-2023-qudeval} evaluated QUDs generated by LLMs by few-shot prompting in a two-step manner: question generation followed by anchor generation. \citet{xu2024graph} followed a QUD style annotation for generating narrative graphs by incorporating retrospective questions triggered from succeeding context. 
 can be put in appendix
\section{The \name{} Framework}
\label{sec:method}

\paragraph{Task Formulation} Given a document with $n$ sentences $D=\{s_1, s_2, \ldots, s_n\}$, QUD parsing aims to build a QUD dependency tree. We formulate the QUD parsing task as edge-level prediction following previous works~\cite{de2018qud, ko-etal-2023-discourse}: given an answer sentence $s_i \in \{s_2, \ldots, s_n\}$\footnote{The first sentence $s_1$ is the root of the QUD dependency tree, and does not anchor on any other sentence}, models are asked to predict the anchor sentence $a_i \in \{s_1, \ldots, s_{i-1}\}$ and generate the question $q_i$. 

\paragraph{Overview} Figure~\ref{fig:method} illustrates the structure of our \name{} framework. We first instruction tune a joint QUD parser \S\ref{sec:qud_train}. Then, we propose selective decoding \S\ref{sec:sel_decode} to select the best candidate from sampled \emph{$\langle$anchor sentence, question$\rangle$} pairs. 


\subsection{QUD Parser Training}
\label{sec:qud_train}
Unlike previous works that use separate models for anchor prediction and question generation, we exploit the instruction following ability of LLMs~\citep{wang2022super} to perform these two steps \textit{jointly}, as demonstrated in Figure~\ref{fig:method}(left). This joint inference provides the model with a holistic view of the task. Given the answer sentence $s_i$ and context of sentences prior to $s_i$, models are instructed to output the anchor  $a_i$ and the question $q_i$. We provide the instruction-response template in Appendix~\ref{app:qud_parser}.


\subsection{Selective Decoding}
\label{sec:sel_decode}
To incorporate specific criteria during inference, we sample multiple \emph{$\langle$anchor sentence, question$\rangle$} candidates and select the best one by using simple criteria scorers.

To generate multiple QUD candidates for a context $\{s_1, \ldots, s_{i-1}\}$ and an answer sentence $s_i$, we sample multiple anchor sentences and question candidates by \textit{selectively} utilizing beam-search with a wide beam while decoding.
First, for anchor prediction, we prompt the model with \emph{sentence $s_i$ is anchored by sentence} using a beam size $k$ to generate $k$ possible anchors.
Post deduplication of anchor candidates, we again utilize beam-search with size $k$ to generate $k$ question candidates for each anchor sentence.
This encourages diversity in both the prediction of anchor sentences and questions. \looseness=-1

We apply $m$ criteria $\mathcal{C}=\{c_1,\ldots,c_m\}$ to assess the quality of generated candidates from different aspects. Each criterion assigns a score $c_j(a, q) \in [0,1]$ to a candidate $\langle a, q\rangle$, and the overall score is the summation of all criteria $\Sigma_{j=1}^m(c_j(a, q))$. The candidate with the highest overall score is selected as the final prediction. 

\paragraph{Criteria Scorers.}
We consider the three key principles of QUD as our criteria: answer-compatibility, givenness, and anchor relevance. We implement \emph{reference-free} and \emph{training-free} scorers for each of them. 

\emph{Answer Compatibility:} This criterion indicates that the question $q$ should be answerable by the answer sentence $s_i$. We regard this as a natural language inference (NLI) task, and use the probability that $s_i$ entails $q$ measured by an off-the-shelf NLI model (\texttt{bart-large-mnli}) as the compatibility score. 

\emph{Givenness:} This criterion evaluates if the question only consists of information from the context. An ideal question should be naturally invoked from the context, without concepts that appear out of thin air. We measure the givenness with content word overlap between $q$ and the context $s_{1\ldots i-1}$. We extract lemmas $L_q$ and $L_c$ of all content words (nouns, verbs, adjectives, and adverbs) in the question and the context, and compute the givenness score as $|L_q \cap L_c|/|L_q|$.

\emph{Anchor Relevance:} This criterion measures if the question $q$ is relevant to the anchor sentence $a$. Similar to the givenness score, we approximate it with content word overlap between $a$ and the focus of $q$. We regard the maximum noun phrase of $q$ as its focus $f_q$, and extract lemmas $L_{fq}$ and $L_a$ of all content words in $f_q$ and $a$. The relevance score is computed as $|L_{fq} \cap L_a|/|L_{fq}|$.


\section{Experimental Setup}
\label{sec:experiments}

\begin{table*}[ht]
\footnotesize
\begin{tabularx}{\textwidth}{Xl}
\toprule
\textbf{\name{} (\mistral{})} & \\ 
\midrule
Answer: $s_3$ Anchor: $s_1$ QUD: ``Why is it important that U.S. exports of nuclear material cannot be adequately traced from country to country?'' & \cmark{Direct answer} \cmark{No new concepts} \cmark{Fully grounded} \\
Answer: $s_4$ Anchor: $s_2$ QUD: ``Who commissioned the report?'' & \cmark{Direct answer} \cmark{No new concepts} \cmark{Fully grounded} \\
\midrule
\textbf{Pipeline (\citet{ko-etal-2023-discourse}}) & \\
\midrule
Answer: $s_3$ Anchor: $s_2$ QUD: ``What does Glenn think is the future outlook on nuclear materials?'' &\xmark{Non answer} \xmark{Answer leakage} \cmark{Partially grounded}  \\
Answer: $s_4$ Anchor: $s_2$ QUD: ``Who is  the Sen. Glenn from?'' & \xmark{Nonsensical question} \\
\bottomrule
\end{tabularx}%
\caption{Example QUDs generated by \name{} (\mistral{}) and the pipeline method for a test article. The full article text can be found in Appendix Figure \ref{tab:case_study_article}. $s_i$ indicates the $i$-th sentence in the article.}
\label{tab:qual_eg}
\end{table*} 

\paragraph{Models and Datasets}
We utilize the DCQA dataset \cite{ko-etal-2022-discourse} for training and evaluating QUD parsers. The DCQA dataset consists of 22k English questions across 606 news articles. We use two instruction-tuned models \llama{}-7B \cite{touvron2023llama} and \mistral{}-7B \cite{jiang2023mistral} as base models of our framework. To explore the effectiveness of selective decoding on closed-source models, we also apply it to GPT-4~\cite{achiam2023gpt}. We sample $k=10$ candidates for each answer sentence. Implementation details can be found in Appendix~\ref{app:qud_parser}. 

\paragraph{Baselines} We compare against two existing QUD parsers: the Pipeline training approach \cite{ko-etal-2023-discourse} and the GPT-4 prompting method \cite{wu-etal-2023-qudeval}. We also provide ablation of not using selective decoding during inference, i.e., \name{} with $k=1$. 

\paragraph{Human Evaluation}
We follow the annotation guidelines outlined in \textsc{QUDeval} and evaluate the quality of the generated QUDs for answer compatibility, givenness, and anchor relevance. Detailed classification of the criteria is in Appendix \ref{app:eval_criteria}. We evaluate 100 questions across 8 articles from the DCQA test set. We recruit three annotators from Amazon's Mechanical Turk (MTurk) after extensive training and qualification studies. We report the majority vote results and achieve an average inter-annotator agreement of 68.3\% averaged across all evaluated dimensions. More details are in Appendix \ref{app:iaa}. 

\paragraph{Automatic Evaluation}
While human evaluation is more accurate for evaluating the efficacy of QUD parsing models, it is time-consuming and expensive to collect at scale. To this end, we apply supervised classifiers to judge the generated QUDs. Specifically, we train RoBERTa classifiers \cite{liu2019roberta} on the expert annotated data in \textsc{QUDeval} for answer compatibility and anchor relevance, and Longformer~\cite{beltagy2020longformer} for givenness due to the longer context length. We achieve a macro F1 score of 0.48 for answer compatibility, 0.42 for givenness, and 0.53 for anchor relevance, outperforming or matching the best existing automatic evaluators. Detailed comparisons with other evaluators are in Appendix \ref{app:auto_eval}.

\section{Results and Analysis}
\label{sec:results}

\subsection{Main Results}

\paragraph{Automatic Evaluation Results.} Table \ref{tab:auto_eval}(top) reports the automatic evaluation results. \name{} (\mistral{}-7B) outperforms the previously established pipeline baseline on all the three criteria. And \name{} improves the performance of instruction tuned \mistral{}-7B, \llama{}-7B and GPT-4, leading to $\sim4\%$ improvement over models without \name{}.

\paragraph{Human Evaluation Results} Table \ref{tab:auto_eval} (bottom) reports the human evaluation results. We compare the best open-source model from Table \ref{tab:auto_eval}, \name{} (\mistral{}-7B), with Pipeline and \mistral{}-7B. \name{} (\mistral{}-7B) generates 67\% directly answered questions, 78\% questions with no unseen concepts, and 68\% fully grounded questions. This highlights the effectiveness of our framework in generating QUDs that satisfy the desired criteria. 

\begin{figure}[th]
    \centering
    \includegraphics[width=\linewidth]{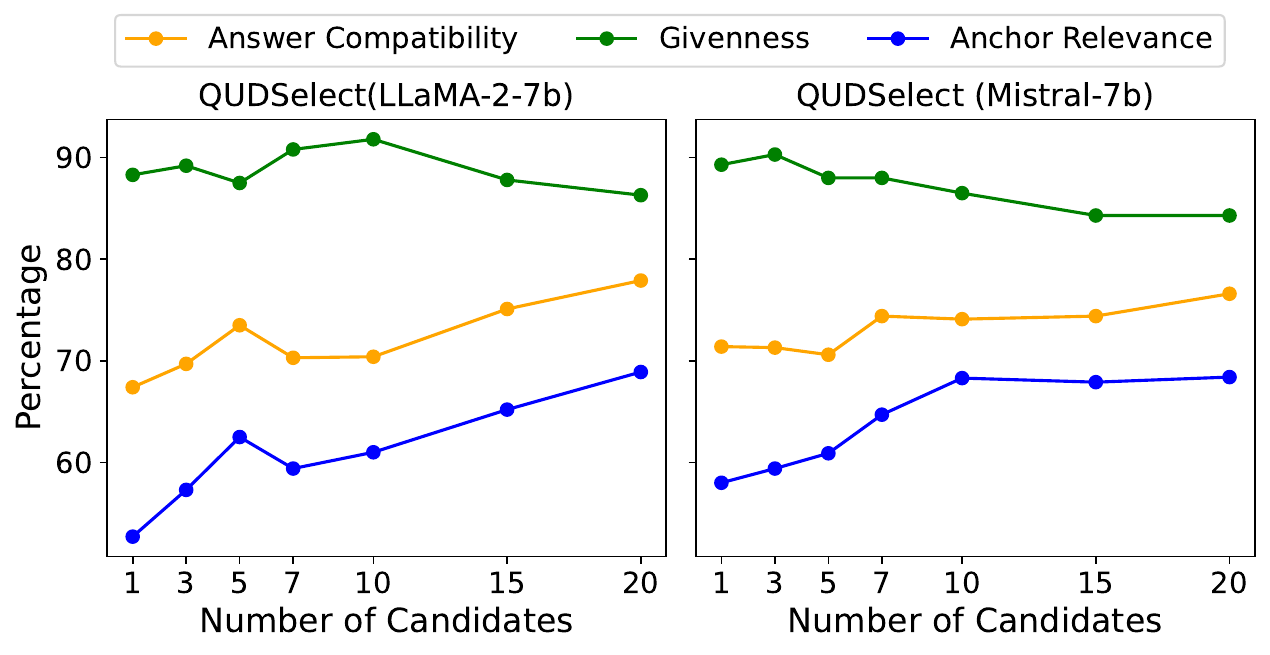}
    \caption{Hyperparameter analysis on the number of candidates. \textsc{QUDSelect} shows improved performance with an increased number of candidates.} 
    \label{fig-hyper}
\end{figure} 

\subsection{Hyperparameter Study}
\label{sec:hyper_study}
To study the performance sensitivity of \name{} to the number of candidates $k$, we vary $k$ from 1 to 20 for \name{} (\llama{}-7B) and \name{} (\mistral{}-7B) and show the performance in Figure \ref{fig-hyper}. 
The performance reveals an upward trend as $k$ grows for Answer Compatibility and Anchor Relevance while Givenness is sacrificed by a small margin for better overall performance. With $k=10$, \name{} significantly outperforms the selected baselines without significant runtime overhead.

\subsection{Case Study}
In Table \ref{tab:qual_eg}, we show the QUDs generated by \name{} (\mistral{}-7B) and the Pipeline model for a news article (Appendix Figure \ref{tab:case_study_article}) along with the human annotations for each question. Most QUDs generated by \name{} (\mistral{}-7B) are explicitly answerable, include no unseen concepts, and are fully grounded in the anchor. In contrast, the Pipeline method generates incomplete questions or incompatible question-answer pairs for the given article. This demonstrates the overall effectiveness of \name{} in generating high-quality QUDs.

\section{Conclusion}
\label{sec:conclusion}
In this work, we propose \name{}, a joint framework for generating QUD structures by integrating key \emph{theoretical criteria}. To achieve this, we reformulate the QUD parsing as an instruction tuning task and selectively decode the candidate questions and anchors.
Furthermore, we develop automated evaluation methods trained on expert annotations to reduce the reliance on labor-intensive expert evaluations and facilitate model development for QUD parsing. Experiments demonstrate that \name{} significantly outperforms baselines in both automatic and human evaluations. 
\pagebreak

\section*{Limitation}
\label{sec:limitation}
\name{} generates the QUD structure as a dependency tree where each sentence is connected to a prior context via a question. This does not guarantee the generation of full, hierarchical QUDs where the answer of a QUD entails the answer of its descendants \cite{roberts2012information}. Furthermore, \name{} generates each QUD edge independently and does not model the relationships between questions. Thus, we leave the exploration of such discourse level constraints to future work. 

\textbf{Sampling Cost.} Although the time cost increases when sampling more candidates for \name{}, the sampled unique anchors tend to be saturated, due to the limited number of reasonable anchors. The average number of unique anchors is still less than 3 when $k=20$. Therefore, the growth of sampling cost is approximately linear to $k$. We find that increasing the number of candidates leads to an increase in the model performance \S \ref{sec:hyper_study}.

\section*{Ethical Consideration}
Our framework relies on open-source and closed-source LLMs that may generate harmful and biased outputs. Therefore, it should be used with human supervision. 
For human evaluation, we recruit annotators
from Amazon Mechanical Turk, and all annotators
are fairly paid more than \$15 USD per hour (it
varies depending on the time spent on HITs), which
is higher than the national minimum wage where
the annotators are recruited.

\section*{Acknowledgements}
We thank Hritik Bansal and Sidi Lu for their constructive comments. We thank the anonymous reviewers for their helpful discussions and suggestions.

\bibliography{anthology,custom}
\clearpage
\appendix

\section{\name{} Implementation Details}
\label{app:qud_parser}
We instruction-tune QUD parsers in the format of Figure~\ref{tab:qud_prompt}. Due to memory limit, we apply LORA (low-rank adaptation,~\citet{hu2021lora}) with learning rate $2e-5$, $lora_{rank}=256$, and $lora_{alpha}=256$. Models are trained for $2$ epochs with batch size $128$. 
During inference, we sample QUD candidates with $k$ beams and temperature $1$. 
All the experiments are performed with 48GB NVIDIA A6000 GPUs.

\begin{figure}[ht]
\centering
\resizebox{\linewidth}{!}{

\begin{tabular}{p{1.3\linewidth}}

\toprule
\#\#\# \textbf{Instruction:} Given the answer sentence, reason through the context to find the most likely sentence where a question can be generated.
\\ \\
\#\#\# \textbf{Input:}\\
Context: \textcolor{airforceblue}{\{context\}} \\
Answer sentence: \textcolor{airforceblue}{\{Answer\}}\\
\\ 
\#\#\# \textbf{Response:} Sentence \textcolor{airforceblue}{\{Answer ID\}} is anchored by sentence \textcolor{airforceblue}{\{Anchor ID\}}, answering the question of ``\textcolor{airforceblue}{\{Question\}}". \\
\bottomrule
\end{tabular}}
\caption{Prompt format for instruction tuning QUD parsers.}
\label{tab:qud_prompt}
\end{figure}

\section{Evaluation Criteria Details}
\label{app:eval_criteria}
We follow the evaluation protocol outlined in \cite{wu-etal-2023-qudeval} for our human and automatic evaluation. 

\begin{itemize}
    \item Answer Compatibility: This criterion indicates that the question $q$ should be answerable by the answer sentence $s_i$. For evaluation, we classify each $q-s_i$ pair as
    a) \textit{Direct and Explicit Answer (Dir.):} $s_i$ answers the $q$ explicitly, b) \textit{Unfocused (Unfocus.):} some parts of $s_i$ answer $q$ indirectly, or c) \textit{Not Answered:} $s_i$ does not answer $q$.
    \item Givenness: This criterion evaluates if the question only consists of information from the context. An ideal question should be naturally evoked from the context, without concepts that are not accessible to the reader from common knowledge. This criterion has the following categories \textit{a) No new concepts (No New):} $q$ does not contain any concepts beyond the context or common knowledge, 
    b) \textit{Answer leakage (Ans. leak.):} $q$ contains concepts that are not in the context but in $s_i$, c) \textit{Hallucination (hall.):} $q$ contains new concepts that are not answer-leakage. 
    \item Anchor Relevance: This criterion measures if the question $q$ is relevant to and naturally evoked from the anchor sentence $a$. This criterion has the following categories \textit{a) Fully Grounded (Fully G.):} $q$ contains concepts from anchor $a$, b) \textit{Partially Grounded (Partial G.):} $q$ contains some concepts from anchor $a$ and is not directly addressing the focus of $a$, c) \textit{Not grounded (No G.)}: $q$ is completely irrelevant to $a$.
\end{itemize}

\section{Human Evaluation Details}
\label{app:iaa}
\begin{figure*}[ht]
    \centering
    \includegraphics[width=\textwidth]{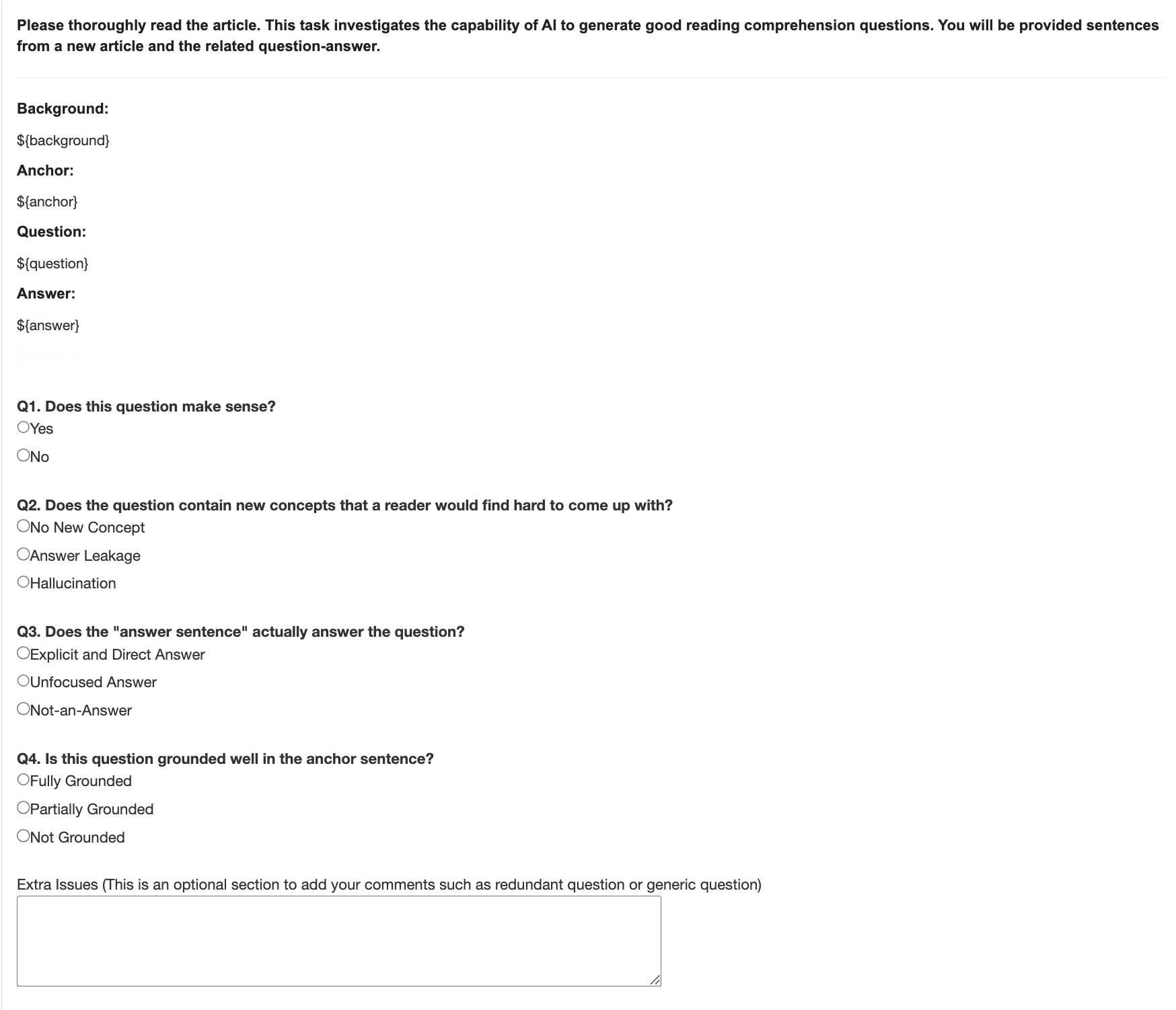}
    \caption{The annotation template for human evaluation. We ask annotators to classify the given QUD, anchor and answer for Givenness, Answer Compatibility, and Anchor Relevance.}
    \label{fig:human_eval_main}
\end{figure*}

\begin{figure*}[ht]
    \centering
    \includegraphics[width=\textwidth]{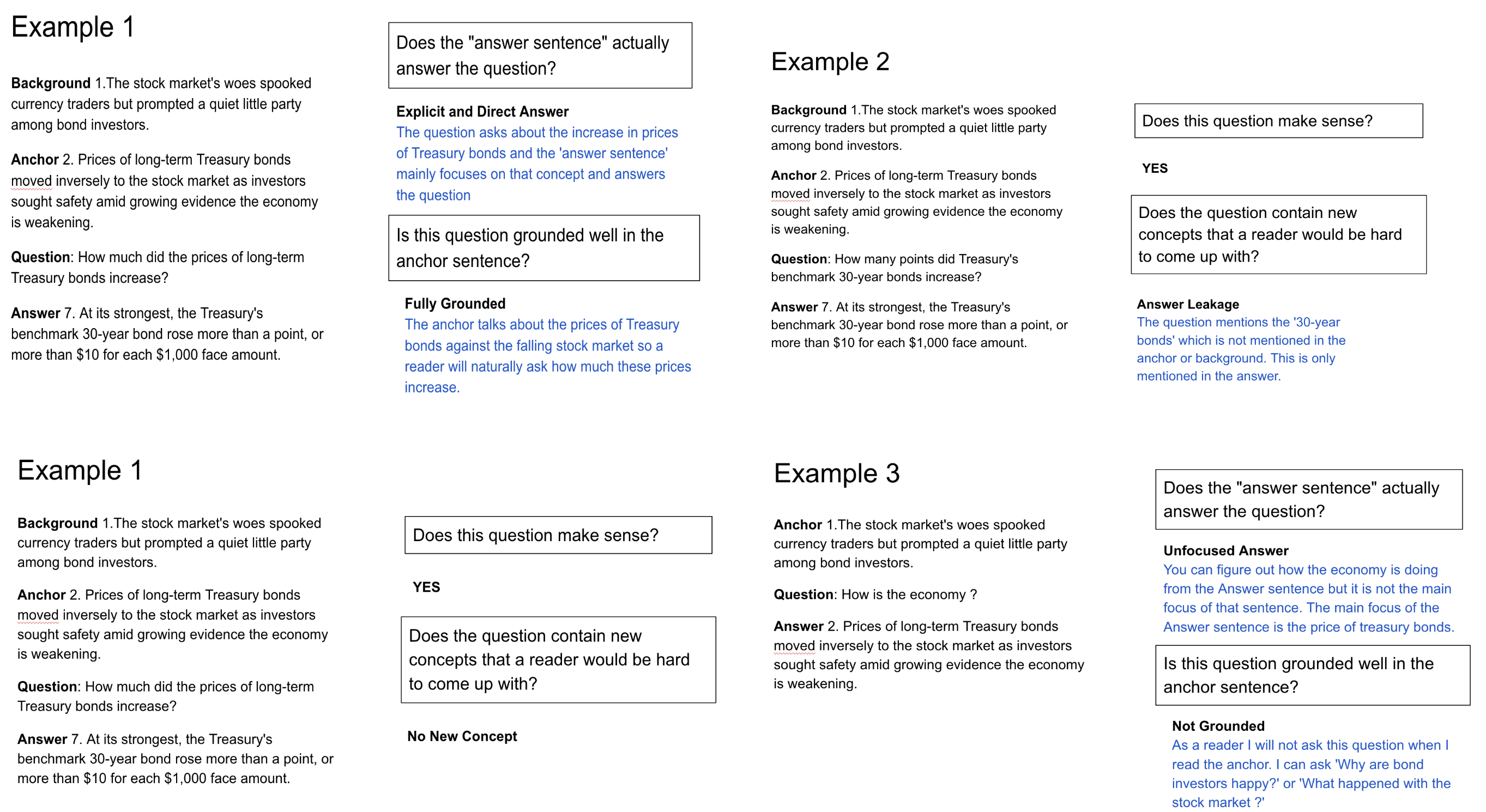}
    \caption{Additional training materials and instructions for human evaluation.}
    \label{fig:human_eval_training}
\end{figure*}

We provide the annotation template and training materials in Figure \ref{fig:human_eval_main} and \ref{fig:human_eval_training}. 

We measure inter-annotator agreement with Krippendorff's $\alpha$. As shown in Table \ref{tab:human_iaa}, annotators achieve ``moderate" agreement across Answer Compatibility and Givenness, while ``fair" agreement for Anchor Relevance~\cite{artstein2008inter}. We also note the pair-wise agreement in Table \ref{tab:human_iaa}. The agreements are comparable with those in \textsc{QUDeval}, and indicate a certain degree of subjectivity in QUD analysis. 

\begin{table}[h]
\centering
\footnotesize
\begin{tabular}{llll}
\hline
& \textbf{Comp.} & \textbf{Givn.} & \textbf{Relv. }\\\hline
Pair-Wise Agreement & 70.0\%  & 75.0\% & 60.0\% \\
Krippendorff's $\alpha$ & 0.68 & 0.64 & 0.43 \\         
\hline
\end{tabular}%
\caption{Inter-annotator agreement for human judges.}
    \label{tab:human_iaa}
\end{table}

\section{Automatic Evaluator Details}
\label{app:auto_eval}
We train automatic evaluators with the human annotations from \textsc{QUDeval}. Experienced human annotators assess the answer compatability, giveness, and anchor relevance of 2,040 machine-generated QUDs from 51 articles. We randomly split the articles into training/validation/test sets with the ratio of $60\%$/$15\%$/$25\%$. 

We fine-tune classifiers for each criterion individually.
We use RoBERTa-large \cite{liu2019roberta} as the backbone model of answer compatibility and anchor relevance, and Longformer-base \cite{beltagy2020longformer} as the backbone model of givenness due to the longer context length. 
For answer compatibility, the input to the model is the question and the answer sentence, and the output is one of the three labels \emph{Dir-Ans.}, \emph{Unfocus.}, and \emph{Not-Ans.} For givenness, the input is the context (sentences before the anchor sentence in the article) and the question, and the answer is one of the three labels \emph{No-New.}, \emph{Ans-leak.}, and \emph{Hallu.} For anchor relevance, the input is the question and the anchor sentence, and the output is one of the three labels \emph{Full.}, \emph{Some.}, and \emph{No-G.} 
Models are fine-tuned for $10$ epochs with the learning rate $1e-5$ and batch size $32$. 

We report the F1 scores of our automatic evaluators in Table \ref{tab:auto_eval_f1}. For reference, we also provide the F1 scores of the random baseline, and the best reference-free and reference-based metrics from \textsc{QUDeval} \cite{wu-etal-2023-qudeval}. GPT-Scr (w/o ref) and GPT-Scr (w/ ref) indicate prompting GPT-4 to score without and with the human-annotated reference QUD. BERTScore means calculating the similarity between the candidate and reference QUD with BERTScore~\cite{zhang2019bertscore}. The rule-based method checks if all content words in the candidate question are presented in the context. Please refer to the \textsc{QUDeval} paper for more details. Note that the results of random and ours are conducted on our held-out test set, while the results of baseline evaluators are conducted on two held-out articles.
Our evaluators are better than or comparable with the baselines, highlighting the credibility of using them in automatic evaluation.

\begin{table}[h]
\resizebox{\linewidth}{!}{
\begin{tabular}{lrrrr}
\toprule
Compatibility & Dir-Ans. & Unfocus.  & Not-Ans. & Macro F1 \\
\midrule
Random        & 0.68  & 0.03  & 0.15  & 0.29  \\
GPT-Scr (w/o ref) & 0.70 &0.05 &0.36 &0.37  \\
BERTScore & 0.51 & 0.14 & 0.43 & 0.36 \\
Ours & \textbf{0.84}  & \textbf{0.28}  & \textbf{0.32} & \textbf{0.48} \\
\toprule
Givenness     & No-New.  & Ans-leak. & Hallu.   & Macro F1 \\ \midrule
Random        & 0.65 & 0.29   & 0.10  & 0.35     \\ 
Rule-based  & 0.52  & \textbf{0.40}   & 0.19      & 0.37        \\
GPT-Scr (w/ ref) & 0.65 & 0.35 & 0.1 &  0.37 \\
Ours & \textbf{0.74}    & 0.23     & \textbf{0.30}   & \textbf{0.42}   \\
\toprule
Relevance     & Full.    & Some.   & No-G.    & Macro F1 \\ \midrule
Random   & 0.52   & 0.22   & 0.21   & 0.32   \\
GPT-Scr (w/o ref)  & 0.73 & \textbf{0.41} & \textbf{0.57} & \textbf{0.57}  \\
GPT-Scr (w/ ref)  & 0.63 & 0.26 & 0.22 & 0.37  \\
Ours & \textbf{0.79}   & 0.32   & 0.48   & 0.53  \\
\bottomrule                 
\end{tabular}%
}
\caption{Automatic evaluator assessment in F1.}
\label{tab:auto_eval_f1}
\end{table}

\section{Article of Case Study}
We provide the article snippet used in the case study in Figure~\ref{tab:case_study_article}. The article is from the DCQA dataset. We also provide questions generated by other models in Table~\ref{tab:more_models}. 

\begin{figure}[ht]
\centering
\resizebox{\linewidth}{!}{

\begin{tabular}{p{1.3\linewidth}}

\toprule
\textbf{1.} U.S. exports of nuclear material cannot be adequately traced from country to country, according to a congressional report.\\
\textbf{2.} 'Scarcely a day goes by without a report of a new black market deal,' said Sen. John Glenn in a statement reacting to the report.\\
\textbf{3.} 'Given the staggering amount of nuclear materials we have exported, it could only be a matter of time before some of this deadly contraband proves to be of U.S. origin.'\\
\textbf{4.} As chairman of the Senate Committee on Governmental Affairs in the last Congress, Glenn commissioned the report from the General Accounting Office, which conducts investigations for legislators.\\
\textbf{5.} The report says hundreds of tons of plutonium and highly enriched uranium have accumulated worldwide, mostly from nuclear power generation.\\
\bottomrule
\end{tabular}}
\caption{Article snippet used in case study.}
\label{tab:case_study_article}
\end{figure}
\begin{table*}[ht]
\footnotesize
\begin{tabularx}{\textwidth}{Xl}
\toprule
\textbf{\llama{}} & \\ 
\midrule
Answer: $s_4$ Anchor: $s_3$ QUD: ``What is deadly contraband?'' & \xmark{Non answer}  \cmark{No new concepts} \xmark{Partially grounded} \\
Answer: $s_3$ Anchor: $s_1$ QUD: ``Why is it difficult to trace nuclear material?"'' & \xmark{Non answer} \cmark{No new concepts} \cmark{Fully grounded} \\
\midrule
\textbf{\name{} (\llama{})} & \\
\midrule
Answer: $s_4$ Anchor: $s_2$ QUD: ``Who requested the report?'' &\cmark{Direct answer} \cmark{No new concepts} \cmark{Fully grounded}  \\
Answer: $s_3$ Anchor: $s_1$ QUD: ``What is the reason for the inability to trace nuclear material?'' & \cmark{Indirect Answer}  \cmark{No new concepts} \xmark{Partially grounded} \\
\midrule
\textbf{GPT4} & \\
\midrule
Answer: $s_6$ Anchor: $s_6$ QUD: ``What does the congressional report reveal about the quantity of nuclear material that has accumulated globally?'' &\xmark{Generated the answer as the anchor and led to answer leakage} \\
Answer: $s_4$ Anchor: $s_2$ QUD: ``Who was responsible for commissioning the report on the traceability of U.S. nuclear material exports?'' & \cmark{No new concepts} \cmark{Fully grounded}\\
\bottomrule
\end{tabularx}%
\caption{Example QUDs generated by different models. The full article text can be found in Appendix Figure \ref{tab:case_study_article}. $s_i$ indicates the $i$-th sentence in the article.}
\label{tab:more_models}
\end{table*}

\end{document}